\documentclass{article}
\usepackage{spconf,amsmath,amssymb,amsfonts,graphicx}
\usepackage{cite,textcomp,xcolor,url,booktabs,multirow}
\usepackage[hidelinks]{hyperref}

\title{Rethinking Correctness for Uncertainty Estimation in Clinical Prediction with Vision--Language Models}
\name{Mingcheng Zhu \qquad Jinning Liang \qquad Tingting Zhu}
\address{University of Oxford, Oxford, United Kingdom\\
mingcheng.zhu@eng.ox.ac.uk}

\begin{document}
\ninept
\maketitle

\begin{abstract}
Vision–language models are increasingly explored for clinical prediction from electronic health records and medical images, where identifying unreliable predictions is important for safe deployment. Uncertainty estimation (UE) enables detecting such predictions, but its evaluation depends on a correctness criterion that determines whether each model output is correct. If this criterion disagrees with human judgement or distorts downstream UE performance, conclusions about model reliability can be misleading. We introduce a two-axis framework that evaluates correctness criteria by their agreement with human judgements and fidelity to human-referenced UE performance. We assess eight criteria across three clinical prediction tasks and three models using 450 predictions annotated by two reviewers. Across the audited tasks, canonical exact matching (EM) achieved the highest observed human agreement and lowest UE distortion, while the BERT-based matching (BEM) and LLM-judge also showed strong human agreement. Across four UE methods and 23,254 clinical predictions, criterion choice changed error-detection AUROC by up to 0.146 and reversed the relative ranking of UE methods. The LLM-judge also selectively accepted invalid or uncertain outputs, accepting 16 of 30 such human-identified errors. These results demonstrate that correctness assessment is an integral component of clinical UE evaluation and should be validated before UE methods are compared. 
Code is available at \url{https://github.com/JasonZuu/EHR-Correctness}.
\end{abstract}


\section{Introduction}
\begin{figure}[ht]
  \centering
  \includegraphics[width=\columnwidth]{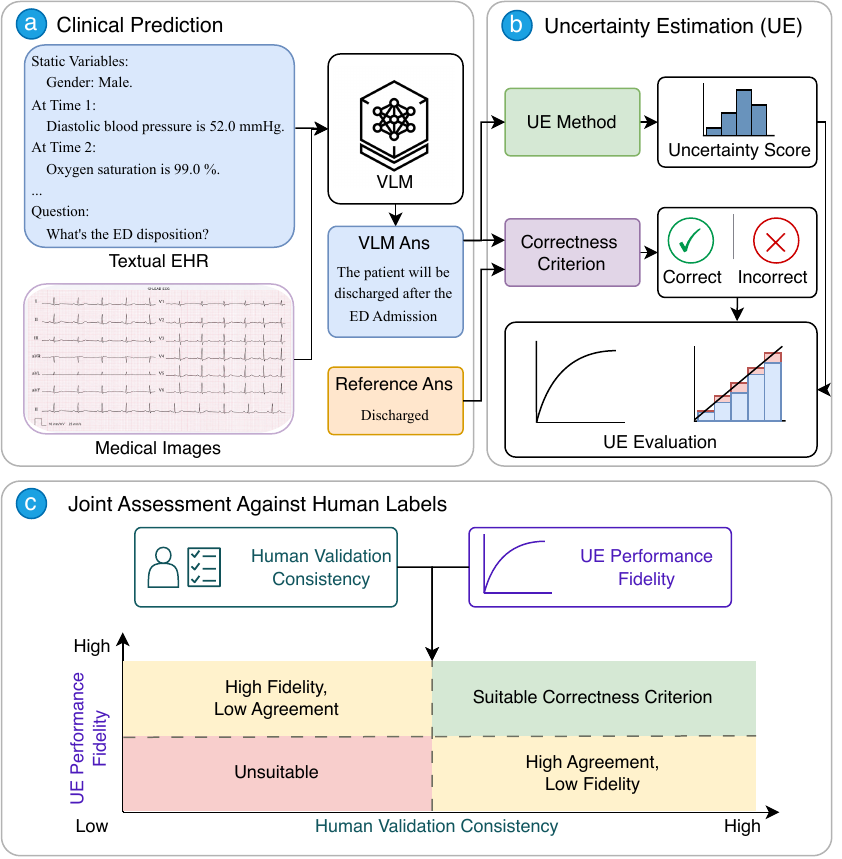}
  \vspace{-10px}
  \caption{Overview of the proposed evaluation framework. (a) A clinical model predicts from an EHR and, where available, an image. (b) An answer-only UE method assigns an uncertainty score, while a correctness criterion compares the prediction with the reference answer to define the binary outcome used for UE evaluation. (c) Joint assessment of agreement with human judgements and preservation of human-referenced UE performance.}
  \label{fig:method-overview}
\end{figure}

Vision–language models (VLMs) are increasingly used for clinical prediction by integrating information from medical images and textual electronic health records (EHRs)~\cite{lou2026key, chen2026cross}. As illustrated in Fig.~\ref{fig:method-overview}(a), these models can leverage multimodal information to support prediction, while reliable assessment of their outputs remains essential for clinical use. Uncertainty estimation (UE) addresses this need by assigning an uncertainty score to each model prediction, as shown in Fig.~\ref{fig:method-overview}(b). However, evaluating UE requires a correctness criterion that determines whether each prediction is correct, and the validity of this criterion is often assumed rather than explicitly validated~\cite{zhu2026towards}. The reliability of UE evaluation therefore depends not only on the uncertainty estimator itself, but also on how prediction correctness is defined.

Automating correctness assessment is challenging because lexical similarity does not necessarily reflect semantic equivalence between model-generated predictions and reference answers~\cite{adlakha2024correctness, farquhar2024semantic}. Several methods have been proposed to provide correctness labels. Canonical exact matching (EM) provides a simple and transparent criterion but may reject valid reformulations, whereas the learned BERT-based matching (BEM) can better accommodate semantic variation~\cite{bulian2022tomayto}. Token-overlap metrics can be unreliable for verbose instruction-following responses~\cite{adlakha2024correctness}. LLM-based judges offer greater semantic flexibility and have shown strong agreement with human assessment in clinical summarisation~\cite{croxford2025clinical}, but broader studies have also identified systematic biases in LLM judging~\cite{chen2024judgement}. Prior work has established that correctness functions can affect UE evaluation \cite{santilli2025revisiting,ielanskyi2026pitfalls}. We instead study correctness criteria themselves as objects of validation, jointly assessing prediction-level human agreement and their downstream distortion of UE evaluation in clinical prediction. Importantly, similar label agreement does not imply similar UE fidelity: two criteria with identical confusion counts may mislabel predictions at different positions in the uncertainty ranking and therefore induce different AUROC distortions. Conversely, preserving aggregate UE performance does not ensure that individual predictions are assessed correctly. We therefore jointly assess correctness criteria through human-label agreement and human-referenced UE performance fidelity. Our contributions are threefold:
\begin{itemize}
\item We provide a consensus-annotated audit of eight correctness criteria across three clinical prediction tasks and three models, using 450 patient-deduplicated predictions.
\item We establish two complementary validity requirements for correctness criteria, agreement with human judgements and preservation of human-referenced UE performance, and operationalise them in a two-axis framework.
\item We quantify criterion-dependent variation in UE estimates on 23,254 parseable predictions and characterise observed false acceptance across clinical response error categories, with explicit scope and uncertainty limitations.
\end{itemize}

\section{Methodology}
\label{sec:methodology}

\subsection{Problem Formulation}
\label{sec:prediction_formulation}

We consider clinical prediction from textual EHRs and, where available, medical images using a VLM. For case $i$, let $x_i=(e_i,v_i)$ denote the textual EHR $e_i$ and medical image $v_i$, with $v_i=\varnothing$ for cases without an image. Given a task prompt $q_t$, a VLM with parameters $\theta$ generates a textual prediction $y_i\sim p_\theta(\cdot\mid x_i,q_t)$.

To estimate the uncertainty of the VLM's generation, a UE method $u$ assigns a scalar score $u_{i}$ to the generated answer $y_i$, with larger values indicating greater uncertainty. Evaluating this score requires a binary outcome indicating whether the prediction is correct. Given a reference answer $\widehat{y}_i$, a correctness criterion $f$ provides this outcome:

\begin{equation} 
c_i=f(y_i,\widehat{y}_i)\in\{0,1\}, 
\label{eq:correctness_label} 
\end{equation}
where $c_i=1$ denotes a correct prediction and $c_i=0$ denotes an incorrect prediction. The resulting correctness defines the reference outcomes used to evaluate whether higher uncertainty is associated with incorrect predictions. Because these outcomes depend on the chosen correctness criterion, different criteria $f$ may yield different estimates of UE performance even when the model predictions and uncertainty scores are the same. We therefore evaluate correctness criteria along two complementary dimensions: agreement with human judgements of prediction correctness and fidelity to UE performance measured using human-derived correctness labels.

\subsection{Human Validation Consistency}
\label{sec:human_consistency}

Human Validation Consistency measures agreement between an automatic correctness criterion and human judgements using balanced accuracy. Let $c_i^{(h)}$ and $c_i^{(f)}$ denote the human and criterion-derived correctness labels, respectively. Let $\mathcal{G}$ contain the $G$ dataset--model groups, and let $\mathcal{I}_g$ contain the human-reviewed predictions in group $g$. We characterise the audited sample using unit analysis weights, $w_i=1$, within each dataset--model group. Within each group, we define
\begin{equation}
\begin{aligned}
H_{f,g}={}&\frac{\sum_{i\in\mathcal{I}_g}w_i c_i^{(h)}c_i^{(f)}}
{2\cdot\sum_{i\in\mathcal{I}_g}w_i c_i^{(h)}}\\
&+\frac{\sum_{i\in\mathcal{I}_g}w_i(1-c_i^{(h)})(1-c_i^{(f)})}
{2\cdot\sum_{i\in\mathcal{I}_g}w_i(1-c_i^{(h)})}.
\end{aligned}
\label{eq:human_consistency}
\end{equation}

The two terms measure agreement on human-labelled correct and incorrect predictions, respectively. We average equally across groups to obtain $H_f=G^{-1}\sum_{g\in\mathcal{G}}H_{f,g}$. Both $H_{f,g}$ and $H_f$ lie in $[0,1]$, with larger values indicating stronger agreement with human judgements. Agreement at the prediction level does not necessarily imply that a criterion preserves the measured performance of UE methods. Balanced accuracy gives equal importance to sensitivity for human-correct responses and specificity for human-incorrect responses within each group. The subsequent macro-average also gives each dataset--model combination equal influence. This separates the summary from differences in group size, while retaining the task and model as the unit of comparison. These weights describe performance on the constructed audit; they do not recover population performance from the EM-stratified sampling design. Consequently, the reported human-agreement estimates should be interpreted together with the sampling protocol and the group-specific results.

\begin{figure*}[th!]
  \centering
  \includegraphics[width=\textwidth]{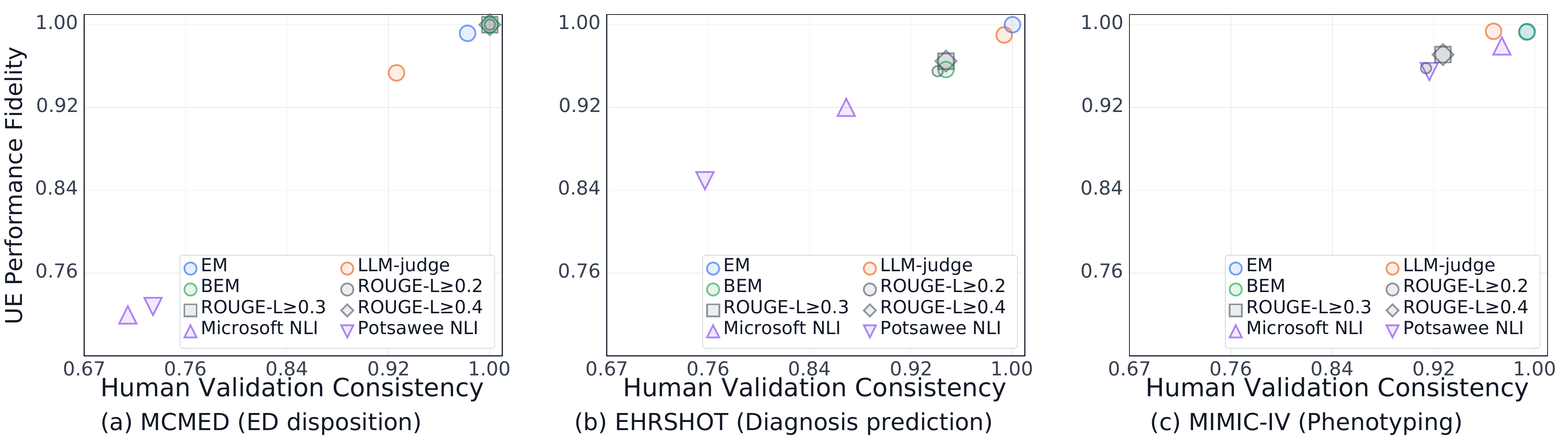}
  \vspace{-15px}
  \caption{Task-specific joint assessment of correctness criteria against the consensus audit. $H$ is balanced accuracy averaged equally across three models; $F$ is one minus absolute human-referenced AUROC distortion averaged across three models and four UE methods. Points use equal case weights within each group. Higher values indicate greater human agreement and UE performance fidelity, respectively.}
  \label{fig:rq1-validity}
\end{figure*}

\subsection{UE Performance Fidelity}
\label{sec:ue_fidelity}

UE Performance Fidelity measures whether a correctness criterion preserves UE performance relative to human-derived correctness labels. For method $u$ and group $g$, let $A_{u,g}^{(f)}$ and $A_{u,g}^{(h)}$ denote AUROCs computed on the same human-reviewed predictions and uncertainty scores. These use criterion-derived error labels $1-c_i^{(f)}$ and human-derived error labels $1-c_i^{(h)}$, respectively, with unit analysis weights. We quantify their difference using the following signed distortion:
\begin{equation}
\Delta_{f,u,g}=A_{u,g}^{(f)}-A_{u,g}^{(h)}.
\label{eq:signed-distortion}
\end{equation}

Positive values indicate higher measured AUROC than the human reference, while negative values indicate lower measured AUROC. Within each group, we average absolute distortion across $N_u$ UE methods and define fidelity as
\begin{equation}
D_{f,g}=\frac{1}{N_u}\sum_{u=1}^{N_u}|\Delta_{f,u,g}|,
\qquad
F_{f,g}=1-D_{f,g}.
\label{eq:mean-distortion}
\end{equation}
We then average equally across groups to get the grouped score $D_f=\frac{1}{G}\sum_{g\in\mathcal{G}}D_{f,g}$ and $F_f=1-D_f$. We report \(F_f\) for visualisation so that both framework axes increase with criterion suitability. Since AUROC lies in $[0,1]$, $F_f\in[0,1]$, with larger values indicating closer agreement with human-referenced AUROCs. A fidelity of one indicates identical AUROCs for the evaluated methods and groups, but does not imply identical correctness labels. Balanced accuracy summarises label agreement but does not specify where disagreements occur in the uncertainty ordering. Because AUROC depends on the score ordering of correct--incorrect pairs, criteria with identical confusion counts can yield different AUROC distortions; conversely, a small mean distortion does not ensure that the underlying correctness labels agree with human judgements.

\subsection{Human Annotation Protocol}
\label{sec:human_annotation}

We construct a human reference from 450 predictions across nine dataset--model groups. Each group contains 50 predictions, stratified equally between EM-accepted and EM-rejected cases to ensure coverage of both canonical-match outcomes rather than estimate their population prevalence. Two trained annotators with clinical-AI research experience independently annotated correctness and a primary error category using the task, candidate labels, reference labels, model answer, reasoning and available model output. Incorrect responses were categorised as missing/partial, extra/wrong or invalid/uncertain; when omissions and additional labels co-occurred, the response was assigned to extra/wrong. All disagreements were resolved jointly. Before discussion, correctness agreement was 445/450 (98.89\%; Cohen's $\kappa$=0.978), and error-category agreement was 206/216 (95.37\%) among responses both annotators judged incorrect. The final reference combines 435 independent agreements and 15 joint decisions, covering five correctness disagreements and ten additional category disagreements. Joint decisions were recorded separately from independent annotations. The consensus reference contains 229 correct responses and 221 incorrect responses across the nine dataset--model groups.

Automatic correctness decisions were hidden during independent annotation. Reviewers assessed whether the answer satisfied the task and reference labels using the information presented in the annotation interface. They did not independently re-evaluate the original clinical records or images. Sampling equal numbers of EM-accepted and EM-rejected predictions ensured that both matching outcomes were examined.

\section{Results}

\subsection{Experimental Setup}

\begingroup\emergencystretch=3em
We evaluate three tasks with predefined candidate-label spaces: prediction of first diagnoses within the following year on EHRSHOT~\cite{wornow2023ehrshot} (six disease labels), ED disposition on MCMED~\cite{mcmed2025} (four outcomes), and ICU phenotyping on MIMIC-IV~\cite{johnson2023mimiciv} (five labels). EHRSHOT and MIMIC-IV permit multiple labels or none; MCMED requires one outcome. MIMIC-IV uses the first 24 hours of ICU events to predict phenotypes at ICU discharge. MCMED uses events recorded before the disposition decision and ECG images; only this task includes image inputs. Models return JSON with reasoning and a string answer naming candidate diagnoses or dispositions. EHRs are represented as textual event streams \cite{zhu2026taxonomies,zhu2026from}. We evaluate Gemma-4 E4B, Gemma-4 26B-A4B and MedGemma-1.5 4B on each task \cite{gemma4report, medgemma15}.
\par\endgroup

We compare eight correctness criteria. Specifically, EM normalises Unicode, case, whitespace and punctuation, maps the complete extracted answer to canonical candidate names, and compares unordered label sets without synonym expansion. Literal ``none'' denotes the empty set. ROUGE-L uses thresholds $\geq$0.20, $\geq$0.30 and $\geq$0.40 \cite{lin2004rouge}; BEM uses $\geq$0.50 \cite{bulian2022tomayto}. Microsoft and Potsawee NLI apply per-label entailment thresholds of $\geq$0.50 followed by exact label-set comparison \cite{he2021deberta,manakul2023selfcheckgpt}. The Gemma-4 31B judge maps the extracted answer to candidate labels without seeing the reference labels; a deterministic step checks set equality and invalid-output flags \cite{gemma4report,zheng2023judging}. The judge receives the task question and candidate labels, but not the source clinical record, image or generation reasoning. Acceptance requires label-set equality and no out-of-space labels or contradictions; malformed judge outputs are not accepted as correct predictions.

We evaluate four answer-only UE methods: Self-Certainty~\cite{kang2025scalable}, LogTokU~\cite{ma2025estimating}, Answer Entropy~\cite{liu2026enhancing} and Semantic Entropy~\cite{farquhar2024semantic}. We assess criterion sensitivity on the full prediction set using error-detection AUROC. This metric quantifies how well uncertainty scores rank incorrect predictions above correct ones, without requiring a decision threshold or a common score scale across UE methods. Our use of AUROC targets a specific question: whether changing the correctness labels changes the apparent ordering quality of a fixed uncertainty score. Predictions and UE scores are held fixed when criteria are compared, so the observed differences arise from the evaluation labels. This analysis does not assess whether uncertainty values are calibrated probabilities, nor does it select an operating threshold for clinical use.

\subsection{RQ1: Human Agreement and UE Fidelity}

This experiment assesses the suitability of correctness criteria through human agreement and UE fidelity. We compare eight criteria on the 450-prediction consensus audit using $H_f$ and $F_f=1-D_f$, with unit case weights and nine equally weighted dataset--model groups. Figure~\ref{fig:rq1-validity} presents the two-axis results by task. We compare BEM and the judge with EM using 10,000 paired bootstrap replicates that preserve each group's EM quotas and share sampled cases across criteria and UE methods. The intervals measure resampling stability within this audit, rather than full-population uncertainty.

EM achieved the highest macro balanced accuracy and lowest mean absolute AUROC distortion, with point estimates of 0.992 and 0.005, respectively. BEM and the judge achieved balanced accuracies of 0.980 and 0.962, with distortions of 0.017 and 0.021. Relative to EM, distortion increased by 0.012 for BEM (95\% paired bootstrap interval: $-$0.001 to 0.026) and 0.016 for the judge (0.004 to 0.037). Using Bonferroni-adjusted 97.5\% CIs for the two comparisons, the judge--EM difference remained positive, whereas the BEM--EM interval crossed zero, so a difference in distortion was not established.

The directional errors further distinguish these criteria. Relative to the human annotations, EM falsely rejected 4/229 correct responses and accepted 0/221 incorrect responses, compared with 0/229 and 9/221 for BEM and 0/229 and 16/221 for the LLM-judge. Among these three criteria, EM exhibited the strictest acceptance behaviour in the audit, rejecting all human-labelled incorrect responses at the cost of rejecting four correct responses. BEM and the judge accepted all correct responses but also accepted some incorrect ones. As EM, BEM and the LLM-judge achieved the highest human agreement and lowest UE distortion among the evaluated criteria, we retain these three criteria for the sensitivity analysis experiment.

\begin{table}[t]
\caption{Criterion-conditioned error-detection AUROC on the full parseable prediction set. }
\vspace{4px}
\label{tab:ue}
\centering
\setlength{\tabcolsep}{2.0pt}
\renewcommand{\arraystretch}{1.03}
\small
\begin{tabular}{clccc}
\toprule
UE Method & Correctness & EHRSHOT & MCMED & MIMIC-IV \\
\midrule
\multirow{3}{*}{\shortstack{Self-\\Certainty}}
& LLM-judge & 0.424 & 0.574 & 0.525 \\
& BEM       & 0.422 & 0.607 & 0.512 \\
& EM        & 0.425 & 0.597 & 0.543 \\
\midrule

\multirow{3}{*}{LogTokU}
& LLM-judge & 0.251 & 0.636 & 0.586 \\
& BEM       & 0.262 & 0.643 & 0.440 \\
& EM        & 0.253 & 0.631 & 0.582 \\
\midrule

\multirow{3}{*}{\shortstack{Answer\\Entropy}}
& LLM-judge & 0.667 & 0.577 & 0.606 \\
& BEM       & 0.639 & 0.584 & 0.518 \\
& EM        & 0.671 & 0.593 & 0.618 \\
\midrule

\multirow{3}{*}{\shortstack{Semantic\\Entropy}}
& LLM-judge & 0.634 & 0.573 & 0.604 \\
& BEM       & 0.611 & 0.582 & 0.522 \\
& EM        & 0.637 & 0.589 & 0.616 \\
\bottomrule
\end{tabular}
\end{table}

\subsection{RQ2: Criterion Sensitivity of UE Evaluation}

This experiment examines how correctness-criterion choice affects the measured performance and relative ranking of UE methods at scale. We evaluate the four UE methods under EM, BEM and the judge on 23,254 predictions. Table~\ref{tab:ue} reports criterion-conditioned error-detection AUROC, with equal weighting of the three evaluated models within each dataset. Criterion choice changes both the magnitude and ordering of measured UE performance. The largest spread among the dataset--method means occurs for LogTokU on MIMIC-IV: AUROC ranges from 0.440 under BEM to 0.586 under the judge, a difference of 0.146. On the same dataset, LogTokU exceeds Self-Certainty under the judge (0.586 versus 0.525), but falls below it under BEM (0.440 versus 0.512). These reversals show that the correctness definition can affect which UE method appears preferable. Higher criterion-conditioned AUROC does not establish closer agreement with human-referenced performance, because human labels are unavailable for the full set. 

These results motivate criterion validation before selecting UE methods for clinical prediction tasks. The spread across criteria is distinct from distortion relative to a human reference. A larger full-set AUROC under one criterion could reflect better error identification or merely a different assignment of correctness labels. Likewise, an AUROC below 0.5 describes the score ordering under the stated error labels and score direction, rather than a criterion-independent property of the UE method.
 
\subsection{RQ3: Observed False-Acceptance Patterns}

\begin{figure}[t]
  \centering
  \includegraphics[width=\columnwidth]{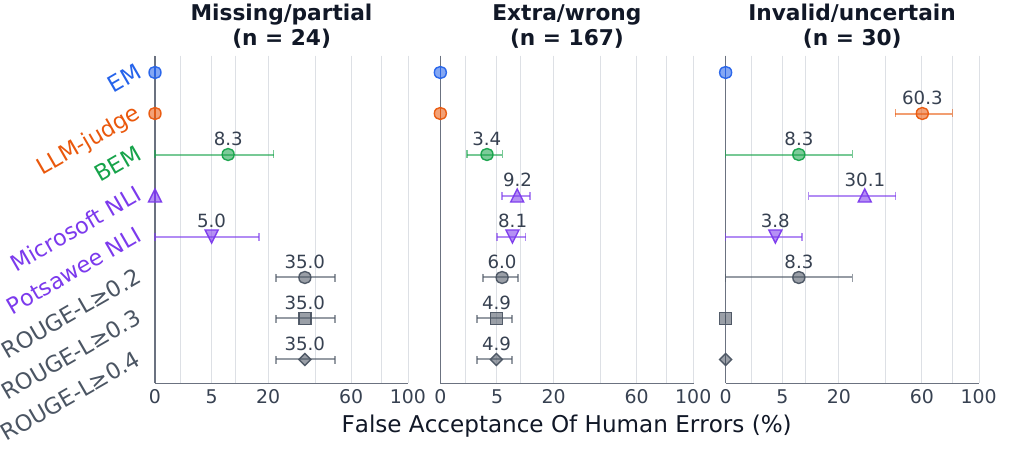}
  \vspace{-22px}
  \caption{Observed false-acceptance rates by human-annotated error category. All groups containing at least one error of the relevant category are included, covering 24/167/30 errors for missing/partial, extra/wrong and invalid/uncertain responses, respectively. Rates are averaged equally across included groups. Error bars show 95\% confidence intervals derived from bootstrap sampling for 10,000 times.}
  \label{fig:rq3-false-acceptance}
\end{figure}

This experiment examines which human-identified clinical errors are incorrectly accepted as correct by each criterion. Figure~\ref{fig:rq3-false-acceptance} reports false-acceptance rates for each category, averaged equally across dataset--model groups containing at least one error of that type. 

Despite its high overall Human Validation Consistency, the evaluated judge accepts 16/30 invalid/uncertain responses, corresponding to an equal-group mean rate of 60.3\%, with no observed false acceptance for missing/partial or extra/wrong predictions. Microsoft NLI also shows its highest false-acceptance rate for invalid/uncertain responses, at 30.1\%, whereas Potsawee NLI shows its highest rate for extra/wrong predictions, at 8.1\%. The three ROUGE-L variants most often accept missing/partial errors, with a mean rate of 35.0\% for this category. BEM shows false acceptance in all three categories, while EM rejects all human-identified errors in the audit. For clinical UE evaluation, an answer's apparent plausibility does not by itself establish that it meets the task's correctness requirements. Accepting invalid or uncertain outputs as correct introduces labels that conflict with human judgements and can either inflate or reduce the evaluated metrics. Human validation should examine incomplete, additional and uncertain answers alongside aggregate agreement and the fidelity of downstream UE estimates.

The concentration of false acceptance in invalid/uncertain responses suggests an error-specific acceptance bias in the evaluated LLM-judge. In these clinical prediction tasks, an answer must provide a determinate response within the candidate-label space. Uncertainty or invalidity cannot be resolved simply by assigning a plausible label. The observed pattern indicates that the judge's acceptance decisions do not consistently preserve this distinction, despite its high aggregate agreement with human annotations. 

\section{Conclusion}

This study shows that reliable evaluation of uncertainty in clinical prediction requires validation of the correctness criteria. Across the evaluated candidate-label tasks, EM achieved the highest observed human agreement and lowest UE distortion. On the full prediction set, criterion choice changed measured AUROC by up to 0.146 and reversed UE-method rankings. Error analysis further revealed the bias of acceptance of invalid or uncertain predictions by LLM-judge. Together, these findings show that correctness assessment is an integral component of clinical UE evaluation and should be validated against both human judgements and downstream UE performance.

\section{Compliance with Ethical Standards}
This study retrospectively analysed de-identified data from MIMIC-IV, MC-MED and EHRSHOT, accessed through PhysioNet and Stanford University under the respective data use agreements. The original MIMIC-IV data sharing was approved by the Beth Israel Deaconess Medical Center Institutional Review Board with a waiver of informed consent. The original MC-MED study was approved by the Stanford University Institutional Review Board, also with a waiver of informed consent. The EHRSHOT study reported that IRB approval was not required because the data were de-identified. Our study involved no patient recruitment or clinical intervention. Human annotation was limited to assessing model-generated responses against predefined task requirements and reference labels.

\section{Acknowledgements}
The authors declare no conflicts of interest.

\bibliographystyle{IEEEbib}
\bibliography{references}

@article{adlakha2024correctness,
  title     = {Evaluating Correctness and Faithfulness of Instruction-Following Models for Question Answering},
  author    = {Adlakha, Vaibhav and BehnamGhader, Parishad and Lu, Xing Han and Meade, Nicholas and Reddy, Siva},
  journal   = {Transactions of the Association for Computational Linguistics},
  volume    = {12},
  pages     = {681--699},
  year      = {2024},
  publisher = {MIT Press},
  doi       = {10.1162/tacl_a_00667},
  url       = {https://aclanthology.org/2024.tacl-1.38/}
}

@inproceedings{bulian2022tomayto,
  title     = {Tomayto, Tomahto. Beyond Token-level Answer Equivalence for Question Answering Evaluation},
  author    = {Bulian, Jannis and Buck, Christian and Gajewski, Wojciech and B{\"o}rschinger, Benjamin and Schuster, Tal},
  booktitle = {Proceedings of the 2022 Conference on Empirical Methods in Natural Language Processing},
  pages     = {291--305},
  year      = {2022},
  address   = {Abu Dhabi, United Arab Emirates},
  publisher = {Association for Computational Linguistics},
  doi       = {10.18653/v1/2022.emnlp-main.20},
  url       = {https://aclanthology.org/2022.emnlp-main.20/}
}

@inproceedings{chen2024judgement,
  title     = {Humans or {LLM}s as the Judge? A Study on Judgement Bias},
  author    = {Chen, Guiming Hardy and Chen, Shunian and Liu, Ziche and Jiang, Feng and Wang, Benyou},
  booktitle = {Proceedings of the 2024 Conference on Empirical Methods in Natural Language Processing},
  pages     = {8301--8327},
  year      = {2024},
  address   = {Miami, Florida, USA},
  publisher = {Association for Computational Linguistics},
  doi       = {10.18653/v1/2024.emnlp-main.474},
  url       = {https://aclanthology.org/2024.emnlp-main.474/}
}

@article{croxford2025clinical,
  title   = {Evaluating Clinical {AI} Summaries with Large Language Models as Judges},
  author  = {Croxford, Emma and Gao, Yanjun and First, Elliot and Pellegrino, Nicholas and Schnier, Miranda and Caskey, John and Oguss, Madeline and Wills, Graham and Chen, Guanhua and Dligach, Dmitriy and Churpek, Matthew M. and Mayampurath, Anoop and Liao, Frank and Goswami, Cherodeep and Wong, Karen K. and Patterson, Brian W. and Afshar, Majid},
  journal = {npj Digital Medicine},
  volume  = {8},
  number  = {1},
  pages   = {640},
  year    = {2025},
  doi     = {10.1038/s41746-025-02005-2},
  url     = {https://doi.org/10.1038/s41746-025-02005-2}
}

@article{farquhar2024semantic,
  title     = {Detecting Hallucinations in Large Language Models Using Semantic Entropy},
  author    = {Farquhar, Sebastian and Kossen, Jannik and Kuhn, Lorenz and Gal, Yarin},
  journal   = {Nature},
  volume    = {630},
  number    = {8017},
  pages     = {625--630},
  year      = {2024},
  publisher = {Nature Publishing Group UK London},
  doi       = {10.1038/s41586-024-07421-0},
  url       = {https://doi.org/10.1038/s41586-024-07421-0}
}

@misc{gemma4report,
  title         = {Gemma 4 Technical Report},
  author        = {{Gemma Team}},
  year          = {2026},
  eprint        = {2607.02770},
  archivePrefix = {arXiv},
  primaryClass  = {cs.CL},
  url           = {https://arxiv.org/abs/2607.02770}
}

@inproceedings{he2021deberta,
  title     = {{DeBERTa}: Decoding-Enhanced {BERT} with Disentangled Attention},
  author    = {He, Pengcheng and Liu, Xiaodong and Gao, Jianfeng and Chen, Weizhu},
  booktitle = {International Conference on Learning Representations},
  year      = {2021},
  url       = {https://openreview.net/forum?id=XPZIaotutsD}
}

@inproceedings{ielanskyi2026pitfalls,
  title     = {Addressing Pitfalls in the Evaluation of Uncertainty Estimation Methods for Natural Language Generation},
  author    = {Ielanskyi, Mykyta and Schweighofer, Kajetan and Aichberger, Lukas and Hochreiter, Sepp},
  booktitle = {International Conference on Learning Representations},
  year      = {2026},
  eprint    = {2510.02279},
  archivePrefix = {arXiv},
  primaryClass  = {cs.LG},
  doi       = {10.48550/arXiv.2510.02279},
  url       = {https://arxiv.org/abs/2510.02279}
}

@article{johnson2023mimiciv,
  title     = {{MIMIC-IV}, a Freely Accessible Electronic Health Record Dataset},
  author    = {Johnson, Alistair E. W. and others},
  journal   = {Scientific Data},
  volume    = {10},
  number    = {1},
  pages     = {1},
  year      = {2023},
  publisher = {Nature Publishing Group UK London},
  doi       = {10.1038/s41597-022-01899-x},
  url       = {https://doi.org/10.1038/s41597-022-01899-x}
}

@inproceedings{lin2004rouge,
  title     = {{ROUGE}: A Package for Automatic Evaluation of Summaries},
  author    = {Lin, Chin-Yew},
  booktitle = {Text Summarization Branches Out},
  pages     = {74--81},
  year      = {2004},
  address   = {Barcelona, Spain},
  publisher = {Association for Computational Linguistics},
  url       = {https://aclanthology.org/W04-1013/}
}

@inproceedings{manakul2023selfcheckgpt,
  title     = {{SelfCheckGPT}: Zero-Resource Black-Box Hallucination Detection for Generative Large Language Models},
  author    = {Manakul, Potsawee and Liusie, Adian and Gales, Mark},
  booktitle = {Proceedings of the 2023 Conference on Empirical Methods in Natural Language Processing},
  pages     = {9004--9017},
  year      = {2023},
  address   = {Singapore},
  publisher = {Association for Computational Linguistics},
  doi       = {10.18653/v1/2023.emnlp-main.557},
  url       = {https://aclanthology.org/2023.emnlp-main.557/}
}

@article{mcmed2025,
  title     = {{MC-MED}, Multimodal Clinical Monitoring in the Emergency Department},
  author    = {Kansal, Aman and Chen, Emma and Jin, Benjamin T. and Rajpurkar, Pranav and Kim, David A.},
  journal   = {Scientific Data},
  volume    = {12},
  number    = {1},
  pages     = {1094},
  year      = {2025},
  publisher = {Nature Publishing Group UK London},
  doi       = {10.1038/s41597-025-05419-5},
  url       = {https://doi.org/10.1038/s41597-025-05419-5}
}

@misc{medgemma15,
  title         = {MedGemma 1.5 Technical Report},
  author        = {Sellergren, Andrew and others},
  year          = {2026},
  eprint        = {2604.05081},
  archivePrefix = {arXiv},
  primaryClass  = {cs.AI},
  url           = {https://arxiv.org/abs/2604.05081}
}

@inproceedings{santilli2025revisiting,
  title     = {Revisiting Uncertainty Quantification Evaluation in Language Models: Spurious Interactions with Response Length Bias Results},
  author    = {Santilli, Andrea and Golinski, Adam and Kirchhof, Michael and Danieli, Federico and Blaas, Arno and Xiong, Miao and Zappella, Luca and Williamson, Sinead},
  booktitle = {Proceedings of the 63rd Annual Meeting of the Association for Computational Linguistics (Volume 2: Short Papers)},
  pages     = {743--759},
  year      = {2025},
  address   = {Vienna, Austria},
  publisher = {Association for Computational Linguistics},
  doi       = {10.18653/v1/2025.acl-short.60},
  url       = {https://aclanthology.org/2025.acl-short.60/}
}

@inproceedings{wornow2023ehrshot,
  title     = {{EHRSHOT}: An {EHR} Benchmark for Few-Shot Evaluation of Foundation Models},
  author    = {Wornow, Michael and Thapa, Rahul and Steinberg, Ethan and Fries, Jason A. and Shah, Nigam H.},
  booktitle = {Advances in Neural Information Processing Systems},
  volume    = {36},
  pages     = {67125--67137},
  year      = {2023},
  publisher = {Curran Associates, Inc.},
  doi       = {10.52202/075280-2933},
  url       = {https://proceedings.neurips.cc/paper_files/paper/2023/hash/d42db1f74df54cb992b3956eb7f15a6f-Abstract-Datasets_and_Benchmarks.html}
}

@inproceedings{zheng2023judging,
  title     = {Judging {LLM}-as-a-Judge with {MT-Bench} and Chatbot Arena},
  author    = {Zheng, Lianmin and Chiang, Wei-Lin and Sheng, Ying and Zhuang, Siyuan and Wu, Zhanghao and Zhuang, Yonghao and Lin, Zi and Li, Zhuohan and Li, Dacheng and Xing, Eric P. and Zhang, Hao and Gonzalez, Joseph E. and Stoica, Ion},
  booktitle = {Advances in Neural Information Processing Systems},
  volume    = {36},
  year      = {2023},
  publisher = {Curran Associates, Inc.},
  doi       = {10.52202/075280-2020},
  url       = {https://proceedings.neurips.cc/paper_files/paper/2023/hash/91f18a1287b398d378ef22505bf41832-Abstract-Datasets_and_Benchmarks.html}
}

@inproceedings{chen2026cross,
  title={Cross-representation benchmarking in time-series electronic health records for clinical outcome prediction},
  author={Chen, Tianyi and Zhu, Mingcheng and Luo, Zhiyao and Zhu, Tingting},
  booktitle={ICASSP 2026-2026 IEEE International Conference on Acoustics, Speech and Signal Processing (ICASSP)},
  pages={7076--7080},
  year={2026},
  organization={IEEE}
}

@article{lou2026key,
  title={Key concept learning for medical vision language model with reasoning capabilities},
  author={Lou, Wei and Wu, Yue and Xu, Pusheng and Zhang, Weiyi and Chen, Xiaolan and Yang, Jiancheng and He, Mingguang and Shi, Danli},
  journal={npj Digital Medicine},
  year={2026},
  publisher={Nature Publishing Group UK London}
}

@article{zhu2026taxonomies,
  title={The Taxonomies, Training, and Applications of Event Stream Modelling for Electronic Health Records},
  author={Zhu, Mingcheng and Liu, Yu and Luo, Zhiyao and Zhu, Tingting},
  journal={arXiv preprint arXiv:2603.14003},
  year={2026}
}

@inproceedings{
zhu2026from,
title={From Token to Token Pair: Efficient Prompt Compression for Large Language Models in Clinical Prediction},
author={Mingcheng Zhu and Zhiyao Luo and Yu Liu and Tingting Zhu},
booktitle={Forty-third International Conference on Machine Learning},
year={2026}
}

@inproceedings{kang2025scalable,
  title={Scalable Best-of-N Selection for Large Language Models via Self-Certainty},
  author={Kang, Zhewei and Zhao, Xuandong and Song, Dawn},
  booktitle={Advances in Neural Information Processing Systems},
  year={2025}
}

@article{ma2025estimating,
  title={Estimating LLM Uncertainty with Evidence},
  author={Ma, Huan and Chen, Jingdong and Zhou, Joey Tianyi and Wang, Guangyu and Zhang, Changqing},
  journal={arXiv preprint arXiv:2502.00290},
  year={2025}
}

@inproceedings{liu2026enhancing,
  title={Enhancing Hallucination Detection through Noise Injection},
  author={Liu, Litian and Pourreza, Reza and Panchal, Sunny and Bhattacharyya, Apratim and Jian, Yubing and Qin, Yao and Memisevic, Roland},
  booktitle={International Conference on Learning Representations},
  year={2026}
}

@article{zhu2026towards,
  title={Towards Generation-Efficient Uncertainty Estimation in Large Language Models},
  author={Zhu, Mingcheng and Liu, Yu and Zhu, Tingting},
  journal={arXiv preprint arXiv:2605.06053},
  year={2026}
}

\end{document}